# Development of a Service Robot for Hospital Environments in Rehabilitation Medicine with LiDAR-Based Simultaneous Localization and Mapping

Sayat Ibrayev, Arman Ibrayeva, Bekzat Amanov, Serik Tolenov
Joldasbekov Institute of Mechanics and Engineering, Almaty, Kazakhstan

*Abstract*— **This paper presents the development and evaluation of a medical service robot equipped with 3D LiDAR and advanced localization capabilities tailored for use in hospital environments. The robot employs LiDAR-based Simultaneous Localization and Mapping (SLAM) to navigate autonomously and interact effectively within complex and dynamic healthcare settings. A comparative analysis with the established 3D-SLAM technology in Autoware version 1.14.0, under a Linux ROS framework, provided a benchmark for evaluating our system's performance. The adaptation of Normal Distribution Transform (NDT) Matching to indoor navigation allowed for precise real-time mapping and enhanced obstacle avoidance capabilities. Empirical validation was conducted through manual maneuvers in various environments, supplemented by ROS simulations to test the system's response to simulated challenges. The findings demonstrate that the robot's integration of 3D LiDAR and NDT Matching significantly improves navigation accuracy and operational reliability in a healthcare context. This study not only highlights the robot's ability to perform essential tasks with high efficiency but also identifies potential areas for further improvement, particularly in sensor performance under diverse environmental conditions. The successful deployment of this technology in a hospital setting illustrates its potential to support medical staff and contribute to patient care, suggesting a promising direction for future research and development in healthcare robotics.**

## I. Introduction

The integration of robotics into healthcare represents a transformative shift in rehabilitation medicine, promising enhanced precision, efficiency, and patient outcomes. Rehabilitation robotics, especially in hospital environments, has seen considerable growth, propelled by advancements in automation and sensor technology. This paper focuses on the development of a medical service robot designed specifically for hospital settings in rehabilitation medicine, employing LiDAR-based Simultaneous Localization and Mapping (SLAM) to navigate and function autonomously [1].
Rehabilitation robots are primarily developed to assist with the delivery of intensive, repetitive, and task-specific interventions which are often labor-intensive and require high levels of precision [2]. The role of these robots extends beyond mere assistance, as they are increasingly equipped with autonomous features that allow them to navigate complex hospital environments and interact with patients and healthcare staff effectively [2]. The adoption of LiDAR technology in medical service robots enhances these capabilities by providing accurate and real-time 3D maps of the environment, which is critical for the autonomous navigation and operational safety of robots [3].
The importance of autonomous navigation systems in medical robots cannot be overstated, as they significantly reduce the human resources needed for operation and maintenance, thereby increasing the healthcare system's overall efficiency [4]. Simultaneous Localization and Mapping (SLAM) technology, which combines data from various sensors to create a map of an unknown environment while simultaneously tracking the robot's location, is pivotal in this context. SLAM has been extensively studied and applied in mobile robotics, and its adaptation to the specific needs of medical environments presents unique challenges and opportunities [5].
The application of SLAM in medical service robots involves not only technical development but also consideration of the ethical, privacy, and safety concerns associated with robotic operations in human-centric environments [6]. Robots in hospitals must adhere to stringent safety standards and be capable of interacting with patients in a manner that complements the therapeutic goals of rehabilitation [7]. Furthermore, the integration of robots into public health settings raises significant privacy concerns, particularly in relation to the storage and handling of sensitive patient data captured by robotic sensors [8].
The development of robots equipped with LiDAR and SLAM for rehabilitation medicine also necessitates a multidisciplinary approach, combining insights from engineering, computer science, and healthcare. Such collaboration is crucial for ensuring that the robots are not only technically proficient but also tailored to meet the practical needs of patients and healthcare providers [9]. Moreover, the implementation of these technologies must be supported by robust clinical trials to validate their efficacy and safety in real-world hospital settings [10].

Past research has demonstrated the potential of robotic aids in enhancing patient engagement and improving recovery outcomes in rehabilitation settings [11]. For instance, robots that assist with walking or deliver physical therapy have been shown to improve mobility and accelerate recovery, providing a level of consistency and repeatability that is difficult to achieve through human intervention alone [12]. The development of a medical service robot with sophisticated navigation and mapping capabilities could further these benefits by enabling more dynamic and responsive interaction with the environment and the patients.

This research aims to bridge the gap between the current capabilities of medical service robots and the evolving demands of modern healthcare facilities. By focusing on the integration of LiDAR-based SLAM technology, the study seeks to address several of the limitations faced by earlier models of rehabilitation robots, such as limited autonomy and the inability to adapt to new and complex environments [13]. The ultimate goal is to develop a robot that not only supports the logistical needs of hospitals but also contributes directly to the therapeutic processes, enhancing the overall quality of care and patient satisfaction [14].

The development of a medical service robot equipped with LiDAR-based SLAM technology for use in rehabilitation medicine represents a significant advancement in the field. This research contributes to a deeper understanding of the technical challenges and clinical implications of deploying autonomous robots in sensitive environments, aiming to maximize both the efficacy and safety of robotic interventions in healthcare settings [15].

## II. Related Works

In the evolving landscape of rehabilitation medicine, the integration of robotics has marked a significant technological advancement, aiming to enhance patient care through automated assistance and precise intervention. The adoption of advanced technologies like LiDAR and Simultaneous Localization and Mapping (SLAM) within medical service robots presents a novel approach to navigating complex hospital environments efficiently. This section reviews the pertinent literature surrounding rehabilitation robotics, with a focus on the incorporation of these sophisticated technologies into their design and functionality. The discussion extends across the technological underpinnings, applications, and the specific challenges faced, thereby setting a foundational context for this research.

*A. Overview of Rehabilitation Robotics.*

Rehabilitation robotics has emerged as a vital tool in modern therapeutic practices, primarily focusing on enhancing patient recovery and automating repetitive therapy tasks. These robotic systems are designed to deliver high-intensity, precise interventions that are essential for the rehabilitation of patients with diverse physical impairments. According to Zhao et al. (2022), rehabilitation robots not only facilitate consistent therapeutic activities but also significantly reduce the physical burden on healthcare providers by automating routine tasks [16].

The evolution of these systems has been marked by significant advancements in their ability to interact with patients and adapt to various therapeutic needs. As highlighted by Hou et al. (2024), the integration of sophisticated sensors and actuators in these robots enables them to perform complex tasks with greater autonomy and accuracy [17]. This technological enhancement improves the quality of interventions and supports a broader range of rehabilitation activities.

Moreover, the clinical impact of rehabilitation robotics is profound, with studies indicating improved patient outcomes in mobility and independence [18]. These robots provide tailored therapeutic exercises that are crucial for effective rehabilitation, making them an indispensable asset in modern healthcare settings.

*B. Technological Foundations in Medical Service Robots.*

Medical service robots incorporate a variety of advanced technologies to enhance their functionality and autonomy in healthcare settings. Central to their operation are automation technologies and intelligent systems that allow these robots to perform a wide range of tasks, from patient care to logistical support within hospitals. According to Avutu et al. (2023), the use of real-time data processing and machine learning enables these robots to make informed decisions and adapt to dynamic environments, significantly enhancing their operational efficiency [19].

Actuators and sensor technologies play pivotal roles in the functionality of medical service robots. These components ensure precise control and interaction capabilities, critical for tasks that require high levels of accuracy such as medication delivery or patient monitoring [20]. Furthermore, the integration of communication interfaces facilitates seamless interaction with healthcare professionals, allowing for efficient coordination and data exchange.

Moreover, the implementation of robotics in medical services often involves complex system architectures that combine hardware and software solutions to meet the stringent safety and performance requirements typical of medical environments [21]. These integrated systems not only ensure patient safety but also contribute to the overall resilience and reliability of robotic operations in healthcare settings.

*C. Use of LiDAR Technology in Robotics.*

Light Detection and Ranging (LiDAR) technology has been pivotal in advancing robotic navigation systems. LiDAR sensors provide accurate distance measurements by illuminating a target with laser light and measuring the reflection with a sensor. This technology's application in robotics, as detailed by Chen et al. (2023), involves creating high-resolution maps of the robot's surroundings, which is essential for navigating through dynamic environments without human input [22]. In medical settings, the precision of LiDAR technology ensures that robots can navigate crowded hospital corridors and interact with patients and staff safely.

*D. Simultaneous Localization and Mapping (SLAM).*

SLAM technology is crucial for autonomous navigation, enabling robots to build a map of an unknown environment while simultaneously tracking their location within it. The convergence of SLAM with medical service robots enhances their operational autonomy. Takanokura et al. (2023) discuss various SLAM algorithms, each with different strengths, catering to the specific needs of the environment and the task at hand [23]. In the context of rehabilitation robotics, the implementation of SLAM allows robots to adapt to new and evolving environments, facilitating seamless integration into hospital settings.

*E. Integration of SLAM in Medical Robotics.*

The integration of Simultaneous Localization and Mapping (SLAM) technology in medical robotics represents a significant advancement in the autonomous operational capabilities of these systems within complex healthcare environments. SLAM technology allows medical robots to dynamically map their surroundings while maintaining an accurate location within the map, which is critical for navigation and task execution in hospital settings [24].

Incorporating SLAM into medical robotics facilitates enhanced spatial awareness and adaptability, enabling these robots to autonomously maneuver through crowded and dynamically changing hospital corridors and rooms. According to Mbunge et al. (2021), the ability to update and refine their environmental models in real-time allows these robots to operate safely and efficiently around both stationary obstacles and moving individuals, such as patients and medical staff [25].

Furthermore, the application of SLAM in medical robotics not only improves operational efficiency but also enhances the interaction capabilities of these robots with their human counterparts. Pereira et al. (2022) highlight that SLAM-equipped robots can more effectively collaborate with healthcare providers, ensuring that therapeutic and logistical tasks are carried out with minimal human intervention [26]. This seamless integration into healthcare workflows greatly contributes to the overall productivity and patient care standards within medical facilities.

*F. Challenges and Limitations.*

Despite the advancements, the integration of LiDAR and SLAM into medical service robots faces significant challenges. Yam et al. (2021) outline several technical challenges, including the high cost of LiDAR sensors and the computational demands of SLAM algorithms, which can limit their widespread adoption [27]. Additionally, ethical concerns regarding patient privacy and data security are paramount, as these technologies often collect sensitive information that could be vulnerable to breaches. Makhdoom et al. (2022) stress the importance of developing robust security protocols to protect patient data and ensure compliance with healthcare regulations [28].

*G. Gap Analysis.*

The current literature reveals several gaps in the application of advanced technologies like LiDAR and SLAM within the domain of medical service robots, particularly in rehabilitation settings. While significant advancements have been made in technical capabilities, there is a lack of comprehensive studies focusing on the practical integration of these technologies in real-world healthcare environments [29]. Additionally, existing research often overlooks the user-centric design and ethical considerations essential for deploying robots in sensitive areas such as patient care [30].

Moreover, despite the potential of these technologies to enhance robotic functionality, there is a notable deficiency in tailored solutions that address specific clinical needs and seamlessly adapt to the unique dynamics of hospital settings [31]. Addressing these gaps through focused research could lead to more effective and contextually appropriate robotic systems that improve patient outcomes and healthcare efficiency [32].

### III. MATERIALS AND METHODS

*A. Data Collection.*

The conceived system of the driven automated guided vehicle (AAGV) was formulated to elevate the execution of manual environmental mapping tasks currently in use through the utilization of self-positioning and autonomous navigation that operates nearby the layout map of the area. Given the fill running of an environmental assessment, the AGV autonomously constructs a map of the surrounding, a process that depends on conventional methods as well. This innovative approach exploits the factory indoor formation for micro-localization and precise mobile robot self-positioning. By using point cloud data as input, 3D Lidar's structural analysis module focuses on the portion gleaned from the wall structures, where the variance is measured subsequently by comparing it with a reference 2D layout map and the mapped horizontal trajectory (or wheel odometry) of the UGV. The filtering method implemented to achieve the localization uses a particle filter with Monte Carlo method, while the base of this navigation is the information determined from the map coordinate and transformed point cloud data [33].

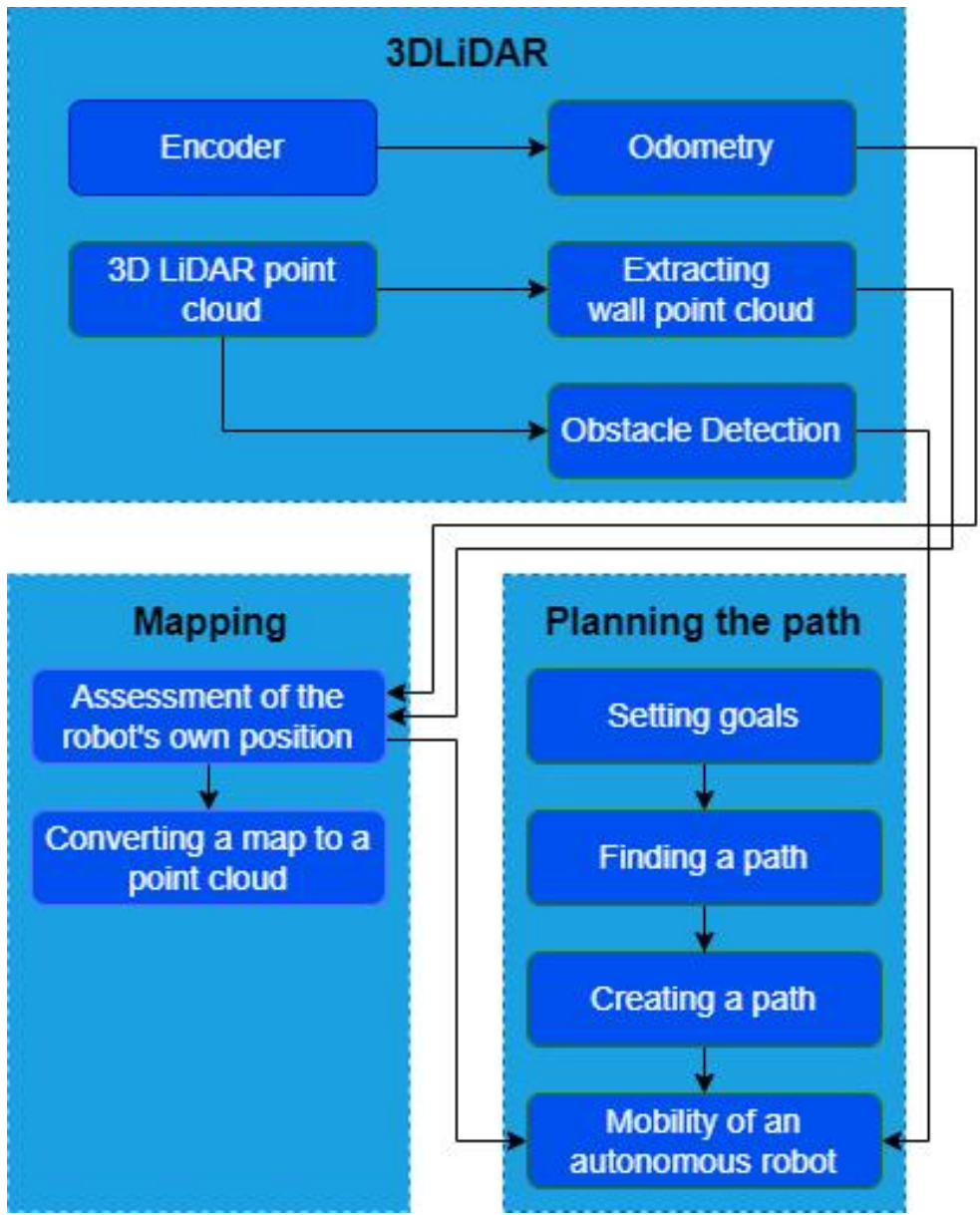

Fig. 1. Basic control scheme of the proposed medical service robot

Figure 1 presents the basic control scheme of the proposed medical service robot, illustrating the integration of 3D LiDAR technology for autonomous navigation within hospital environments. The control architecture is divided into three primary components: 3D LiDAR processing, Mapping, and Path Planning. Initially, the 3D LiDAR sensor collects point cloud data which is processed by the Encoder and Odometry to track the robot's position and movement. Wall point clouds are extracted to delineate boundaries and detect obstacles, ensuring the robot avoids collisions. Subsequently, the Mapping process involves assessing the robot's position and converting the environmental map to a point cloud format for real-time navigation updates. In the Path Planning segment, the robot sets navigation goals, computes feasible paths, and creates a navigational trajectory, culminating in the autonomous mobility of the robot. This control scheme underscores a comprehensive approach to navigating complex healthcare settings, leveraging advanced sensing and computational techniques to enhance operational efficacy and safety.

*B. The Hardware Module.*

In this study, the Shenzhen Yahboom Technology Rosmaster X3 Plus mobile robot was utilized as a primary research tool. This robot, operating within a hybrid system environment, combines physical hardware with virtual systems managed through Ubuntu 20.04 on a VMware Workstation virtual machine, and further controlled via the ROS-Noetic operating system specifically tailored for robotic management. The choice of the Rosmaster X3 Plus for this research is predicated on its advanced capabilities and adaptability to complex tasks, making it an ideal candidate for detailed study in robotic navigation and interaction within structured environments.

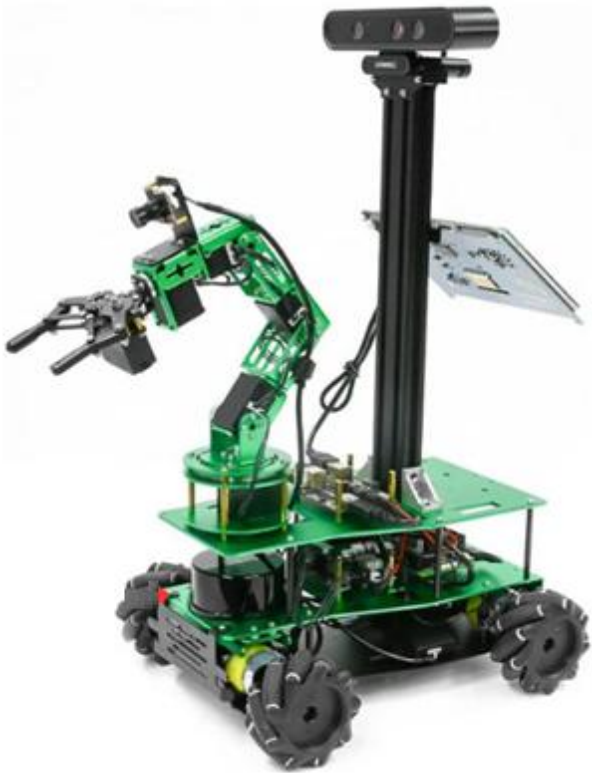

Fig. 2. Medical service robot of the study

The Rosmaster X3 Plus is equipped with cutting-edge hardware that enhances its sensing and computational abilities, crucial for effective navigation and task execution. Central to its operation is the Jetson Orin NX processor, boasting 16GB of memory, which facilitates robust real-time processing for tasks such as obstacle navigation and localization. Environmental perception is significantly enhanced by the integration of a YDLidar 4ROS Lidar system, which provides high-resolution 3D point cloud data. Complementing this, an Astra Pro depth camera provides detailed 3D visual inputs that are integrated with the Lidar data for a comprehensive environmental understanding. This hardware-software synergy not only boosts the robot's operational efficiency but also underscores the effectiveness of modern technologies in autonomous robotic navigation.

The operation of the motor is governed by signals originating from Pulse Width Modulation (PWM), which dictate the motor's speed, and directional signals that guide the rotation. These commands are dispatched by a microcomputer integrated within the motor driver, initiating motion based on inputs.

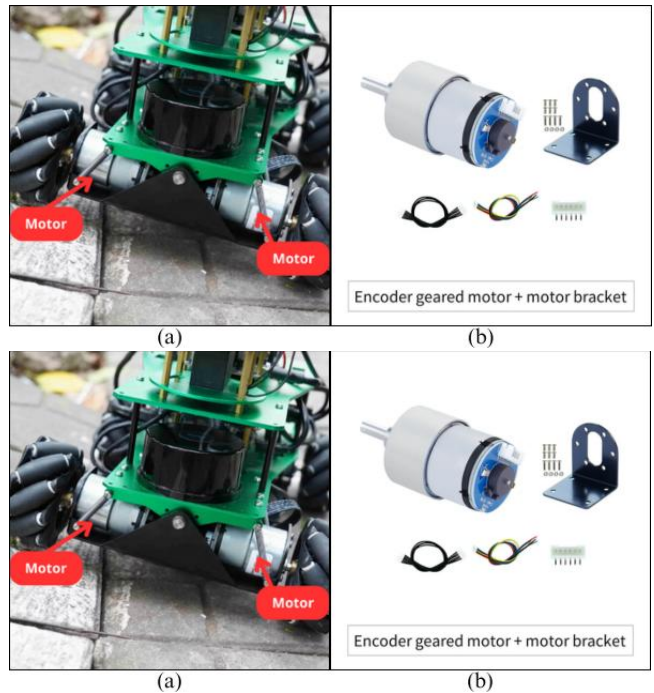

Fig. 3. Motor drive setup (a). DC Motor setup. (b) Encoder configuration.

Further, a serial communication line facilitates the transmission of these commands from the microcomputer to a personal computer (PC), allowing operators to manage the cart's functions effectively through PC-based controls. Additionally, communication between the motor's encoder and the microcomputer is handled via an SPI interface, which transmits precise rotational angle data, subsequently relayed to the PC, enhancing clarity in monitoring and controlling the motor's activity and the cart's trajectory.

*C. Odometry.*

The integration of 3D LiDAR with wheel odometry equips the robot with enhanced capabilities to both perceive its surroundings and track its movement accurately. Odometry, fundamentally reliant on mathematical equations and principles, plays a crucial role in this process. It operates by analyzing the rotation of the robot's wheels, which directly informs the calculation of the distance traveled. Each wheel is equipped with an encoder that records the number of rotations, allowing for precise measurement. Given the radius $r$ of the wheels and the number of encoder ticks $N$, the distance $D$ each wheel travels can be calculated, providing critical data for navigating and mapping the robot's environment effectively

$$D = 2\pi r \, \frac{N}{N_{total}} \quad (1)$$

Where $N_{total}$ is the total number of ticks per complete wheel rotation.

For a two-wheeled robot, the distance traveled $(D_{avg})$ and change in orientation are given by:

$$D_{avg} = \frac{D_L + D_R}{2} \quad (2)$$

$$\Delta\theta = \frac{D_R - D_L}{W} \quad (3)$$

Where $D_L$ and $D_R$ are the distances traveled by the left and right wheels, respectively, and $W$ is the width between the wheels.

3D LiDAR technology generates a detailed point cloud that captures the robot's surroundings, facilitating a comprehensive understanding of its movements and orientation through external reference points. When combined with wheel odometry, LiDAR helps to correct potential inaccuracies and drifts that may accumulate in the odometry data over time.

The process of merging odometry with 3D LiDAR data involves a methodical approach:

1. Initial Estimation: Wheel odometry is initially used to estimate the robot's trajectory.

2. LiDAR Correction: The point cloud produced by LiDAR is compared against a pre-established map to identify any deviations that suggest errors in the odometry data.

3. Data Fusion: Techniques such as the Kalman filter are employed to amalgamate the data from both odometry and LiDAR. This integration enhances the accuracy of the robot's navigation system by providing a more reliable data set that accounts for any discrepancies identified between the odometry and LiDAR inputs.

$$\hat{x}_{k|k} = \hat{x}_{k|k-1} + K_k (y_k - H\hat{x}_{k|k-1}) \quad (4)$$

Where $\hat{x}_{k|k}$ is the a posteriori state estimate, $\hat{x}_{k|k-1}$ is the a priori estimate, $K_k$ is the Kalman gain, $y_k$ represents the measurement (Lidar data), and $H$ is the measurement matrix relating the state to the measurement.

Update Position and Orientation: Adjust the robot's estimated position and orientation based on the fused data:

$$\begin{aligned} x' &= x + \Delta x \\ y' &= y + \Delta y \\ \theta' &= \theta + \Delta\theta \end{aligned} \quad (5)$$

Subsequently, the robot's position and orientation are updated based on corrected data derived from the integration of wheel odometry and LiDAR measurements. This fusion ensures that the robot accurately maintains its location and direction, reducing errors that might arise from wheel slippage or uneven terrain.

By combining 3D LiDAR data with wheel odometry, our mobile robot achieves superior navigation precision and detailed environmental mapping. This sophisticated odometry system underpins the robot's ability to autonomously operate in complex and dynamically changing environments, facilitating reliable performance across various operational scenarios.

*D. Localization.*

The navigation capability of our mobile robot is enhanced by 3D LiDAR technology, utilizing a sophisticated localization algorithm that precisely determines the robot's position within its operating environment. This section delves into the mathematical principles and operational mechanics of the localization algorithm employed in our system, emphasizing the integration of 3D scanning data to improve node localization accuracy.

Central to our localization strategy is Monte Carlo Localization (MCL), also known as particle filter localization. This probabilistic method uses a collection of hypothetical particles to represent potential positions and orientations (states) of the robot within its environment. Each particle is weighted based on its congruence with environmental data gathered via LiDAR scans and the robot's observed movements, effectively merging sensor inputs with motion data to estimate the robot's location with higher accuracy.

Particle Representation: In our system, each particle in the set represents a potential state of the robot, encompassing both its location and orientation, forming the basis for calculating the most probable actual state of the robot as it navigates.

$$p_i = (x_i, y_i, \theta_i) \quad (6)$$

$x_i, y_i, \theta_i$ denote the particle's position and orientation.

Weight Calculation. The weight $w_i$ is calculated for each particle taking into account the degree of matching of the predicted sensor readings for the desired particle state with the real sensor readings given by the 3D Lidar.

$$w_i = P(z_t | p_i) \quad (7)$$

Where $z_t$ is the Lidar measurement at time $t$, and $P(z_t | p_i)$ is the likelihood of observing $z_t$ given the state represented by particle $i$.

Resampling within the Monte Carlo Localization (MCL) framework involves selecting particles based on their weights, with particles possessing higher weights more likely to be chosen. This process concentrates the particle distribution around the most likely states of the robot's position, refining the model's accuracy over time.

Integrating 3D LiDAR data significantly enhances the MCL algorithm's localization precision by allowing a detailed comparison between the environmental features detected by the LiDAR and the pre-existing map model. The LiDAR's point cloud captures environmental details at a granular level, facilitating highly accurate weight calculations for each particle within the model.

The sensor model associated with the 3D LiDAR converts the point cloud data from a sequential format into a probabilistic one, aligning it with the map's specifications. This transformation allows for an effective comparison, essentially converting 3D data into a more manageable 2D format to match the map, thereby enhancing the fidelity and utility of the particle data.

Before assigning weights to each particle, a motion update is conducted based on the reported movements of the robot. This update adjusts the positions and orientations of the particles to reflect the robot's dynamics as captured by its odometry data, ensuring that the model remains consistent with the robot's actual movements. The motion model updates are critical for maintaining the accuracy of the localization process.

$$p_i^{'} = p_i + \Delta p(u_t, ) \quad (8)$$

where $p_i^{'}$ is the updated particle state, $\Delta p$ is the change in state due to the control input $u_t$ (e.g., velocity, angular velocity) at time $t$, and represents the motion noise.

### E. Obstacle Avoidance.

Obstacle avoidance is a critical component of autonomous navigation systems, involving two main functions: detecting obstacles and formulating alternate paths to circumvent them. Consider a scenario where a predetermined route is blocked on the left side by an obstruction. In such cases, navigational coordinates are structured into waypoints, each associated with a specific detection zone. This zone is typically envisioned as a cylindrical area encompassing each waypoint along the robot's path, serving as a detection field for obstacles.

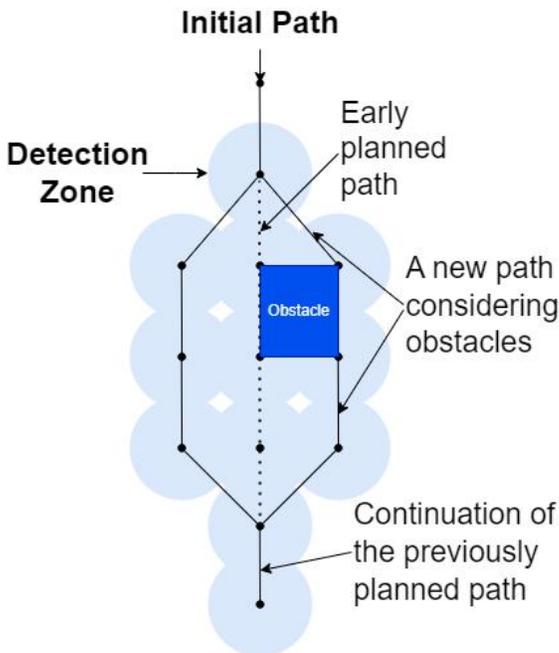

Fig. 4. Obstacle avoidance of the proposed medical service robot.

Within this cylindrical detection zone, any detected objects that do not correspond to the ground are classified as obstacles. The presence of these obstacles renders the area impassable, necessitating a rerouting of the planned path. This mechanism ensures that the robot can adapt its route in real-time to avoid obstacles, maintaining smooth and continuous navigation as depicted in Figure 4.

The obstacle avoidance strategy within autonomous navigation systems functions by designating areas where obstacles are detected as blocked, typically highlighted in red on navigational maps. This prompts the system to seek alternative corridors for maneuvering around the impediment. The algorithm evaluates possible detours, actively searching for viable paths adjacent to the obstruction. If a feasible route is identified along one side of the obstacle, it is selected for navigation, and the route-finding algorithms are updated to reflect this new path. Conversely, if obstructions block all potential routes, rendering them impassable, the vehicle halts its progress until an alternate path becomes available. This adaptive mechanism ensures that the vehicle can flexibly and effectively navigate through varying environmental conditions by dynamically adjusting its course in response to encountered obstacles.

## IV. RESULTS

This section explores the evaluation of the map accuracy generated by our medical service robot, which is crucial for self-localization and overall system performance assessment. Our system was rigorously tested against the renowned 3D-SLAM technology implemented in Autoware version 1.14.0, an established open-source platform designed for autonomous driving technologies under the Linux ROS framework. A critical aspect of Autoware's capability is the Normal Distribution Transform (NDT) Matching, which utilizes point cloud scan matching to enhance localization accuracy. This method employs normal distributions to model point clouds within specified segments, facilitating precise alignment of overlapping point clouds, a feature vital for accurate localization in environments requiring high precision, such as autonomous navigation systems.

To empirically validate our system, data collection involved manually maneuvering a mobile trolley through various environments, recording its positions to verify the accuracy of the self-localization predictions. This testing covered diverse measurement points, extending through indoor and outdoor settings and spaces between different structures. Additionally, our self-localization methods were tested through ROS simulations using the collected positional data. This detailed validation approach ensures that our system's performance is thoroughly understood and reliable in practical scenarios, demonstrating robust capabilities in a real-world application context.

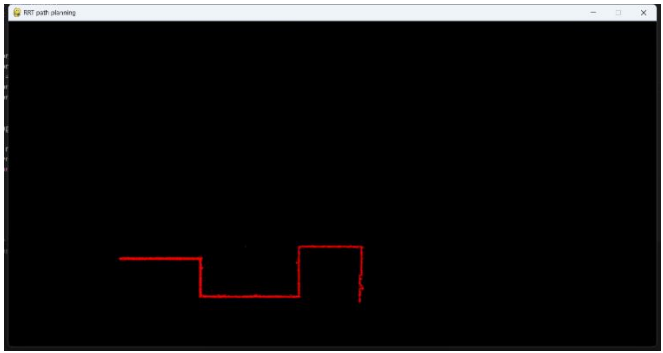

*a) Path planning from lef to wright side*

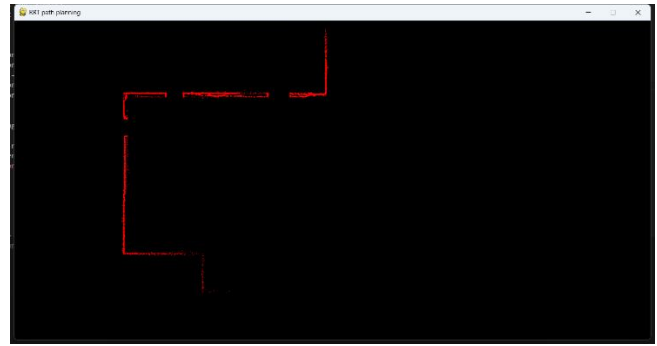

*b) Path planning from bottom to up*

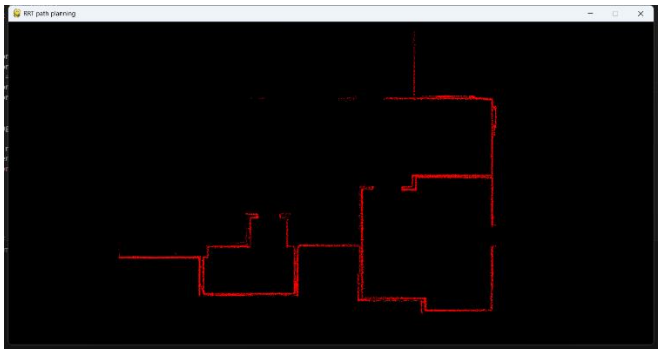

*c) Path planning process*

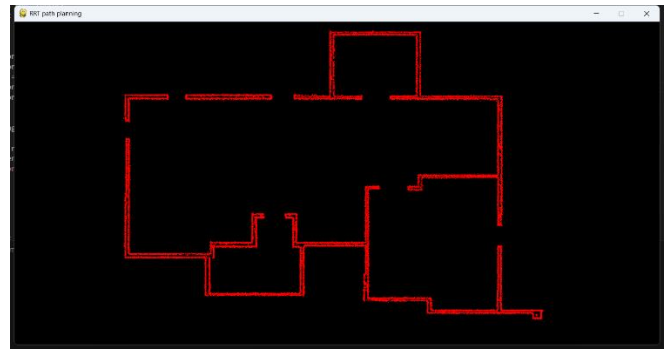

*d) Path planning process has been completed*

Fig. 5. Path planning process for the medical service robot

Figure 5 presents a series of diagrammatic representations illustrating the navigation process within a two-dimensional environment, capturing the dynamic nature of path planning. The sequence starts with Figures 5a and 5b, which mark the initiation of the path planning phase and lay the groundwork for subsequent navigational decisions. This is followed by Figure 5c, which details the iterative steps involved in path planning as the robot maneuvers through various directions. This stage highlights the dynamic and repetitive nature of adjusting the planned route as the robot encounters different scenarios. Conclusively, Figure 5d captures the culmination of the path planning process, displaying a finalized map that delineates the actual path taken by the robot. This sequence effectively demonstrates the progression from initial path determination through to adaptive adjustments and the final mapping, underscoring the complex and responsive strategy employed in robotic navigation.

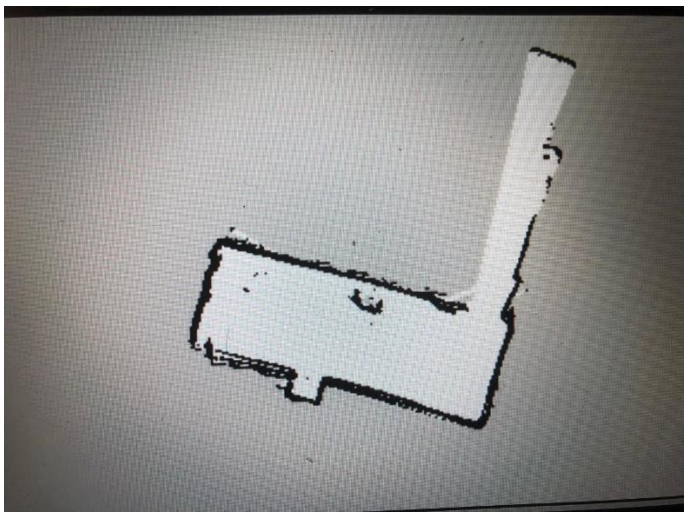

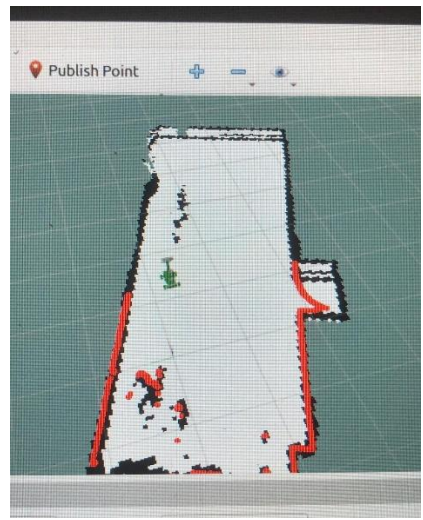

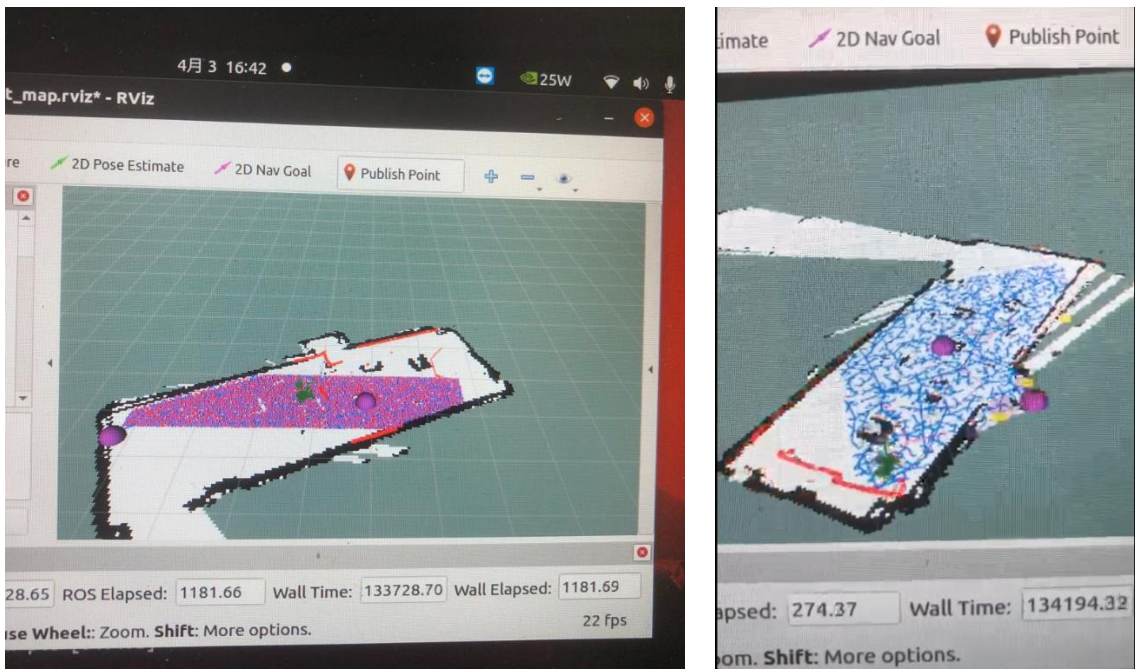

Fig. 6. Path planning in 3D for medical service robot from various foreshortening

Figure 6 provides a detailed exploration of the path planning processes implemented by a robot within a three-dimensional framework, showcasing the steering strategies employed as it navigates through cluttered environments. This diagram effectively illustrates the robot's capability to assess and adapt its trajectory in real-time as it maneuvers through various terrains and obstacles in 3D space. It serves as a critical visual tool for understanding the sophisticated 3D capabilities of autonomous systems, highlighting their ability to perceive and interact with their surroundings comprehensively. The figure emphasizes the advanced algorithms that enable these systems to not only navigate intelligently but also to create detailed 3D maps upon the completion of their routes. This depiction confirms the complexity and dynamism of new 3D path planning techniques, reflecting significant advancements in the autonomous vehicle industry, where precision and efficiency are paramount in navigating complex environments.

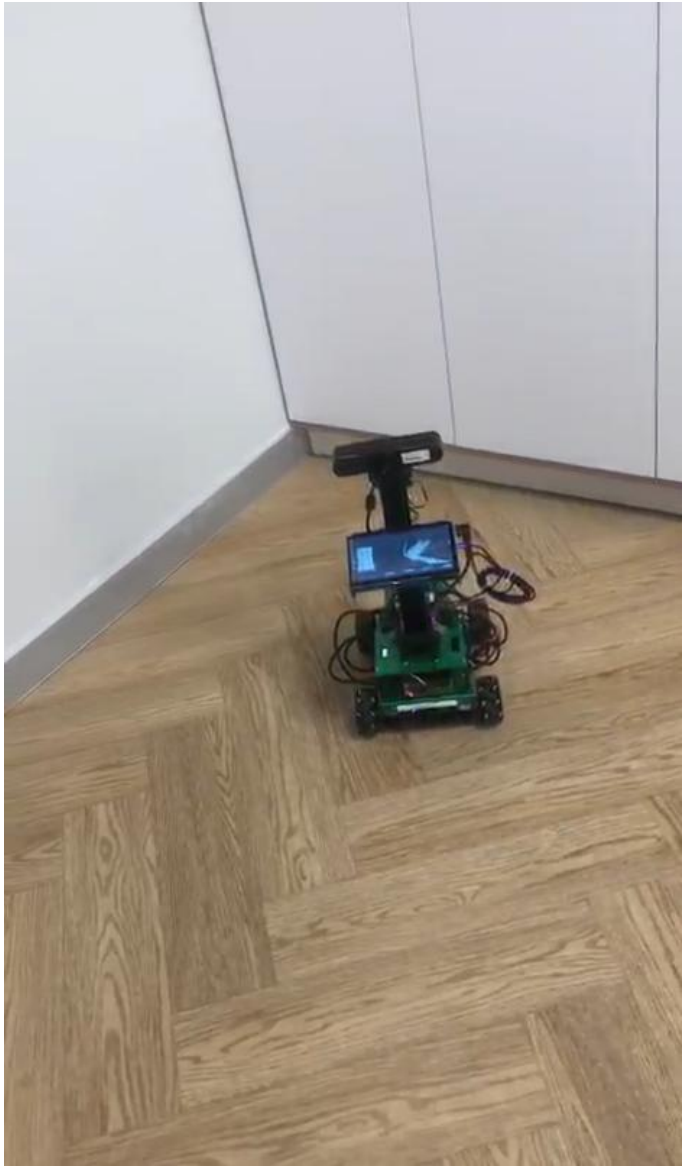

*a) Strat of mobile robot path planning*

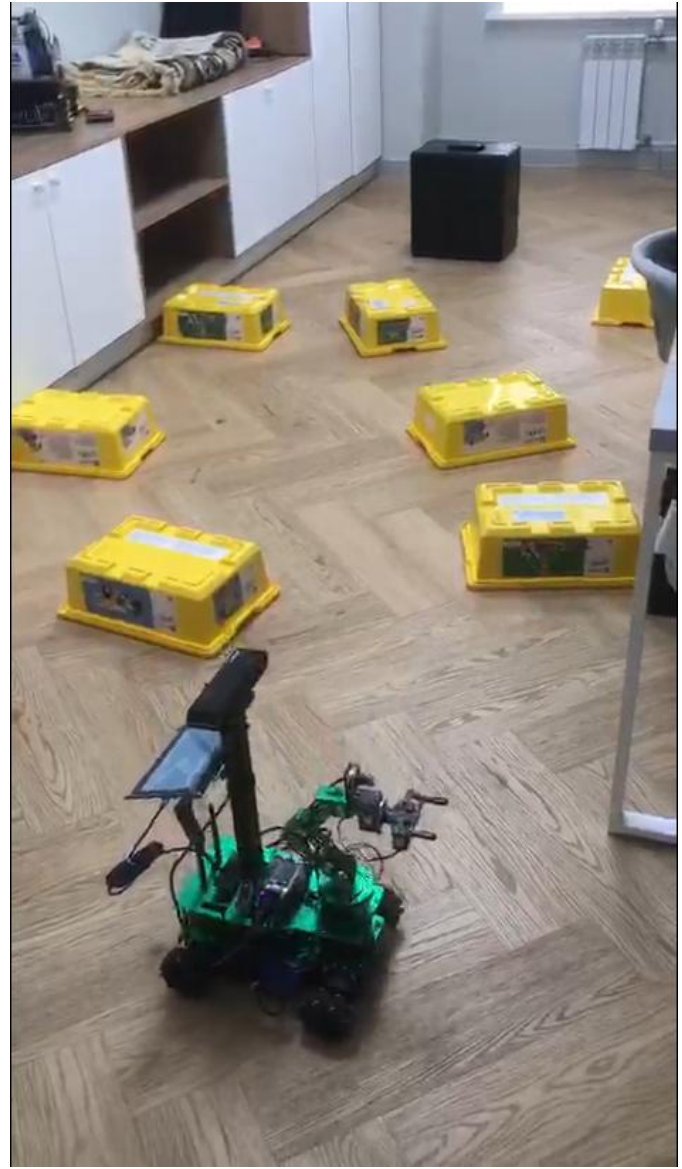

*b) Mobile robot navigation*

Fig. 7.  Mobile medical service robot navigation using 3D LiDAR

Figure 7 showcases the developed mobile robot equipped with 3D LiDAR technology, operating in a real-world or field setting, which illustrates the robot's design and its operational capabilities. This image particularly highlights how the integrated 3D LiDAR technology enables the robot to perceive and interact dynamically with its environment. The diagram captures the robot as it navigates a specified area, utilizing LiDAR data to facilitate steering, obstacle avoidance, and localization tasks. Through this visual representation, the practical functioning of the mobile robot is conveyed, underscoring the real-time application and demonstrating the effectiveness of the combined technologies in enhancing the navigation system. This figure serves as a vital link between the theoretical concepts underlying the study and their practical implementation, showcasing the translation of academic research into actionable, autonomy-enhancing strategies within robotic systems.

V. DISCUSSION

In this discussion, we delve into the findings from the deployment and validation of our medical service robot, equipped with advanced 3D LiDAR technology and an innovative localization system. The results underscore the robot's potential to revolutionize navigation and interaction within hospital environments, providing critical insights into both the capabilities and areas for further enhancement of autonomous robotic systems in healthcare settings.

The application of 3D LiDAR technology in our medical service robot has proven to be a cornerstone for enhancing autonomous navigation. The high-resolution data obtained from LiDAR not only facilitated accurate real-time mapping but also significantly improved the robot's ability to detect and navigate around obstacles. This capability is critical in a hospital setting where dynamic obstacles such as moving people and medical equipment are common. The integration of 3D LiDAR with the robot's other sensory systems has enabled a level of situational awareness that is paramount for safe and efficient operation within such complex environments.

Our comparative analysis with Autoware's 3D-SLAM technology highlighted the effectiveness of our localization approach. The Normal Distribution Transform (NDT) Matching method, typically used in autonomous vehicular navigation, was adapted for indoor use with our robot. This adaptation was crucial as it addressed the unique challenges of indoor navigation, which include lower GPS reliability and the presence of numerous static and dynamic obstacles. The successful application of NDT in our system underscores its potential for broader application in other robotic systems that operate in similarly challenging environments.

Furthermore, the empirical validation of our robot's localization accuracy through manual maneuvering across different environments provided substantial evidence of its robustness. The robot demonstrated a high degree of precision in maintaining its course within tightly controlled trajectories, an essential feature for medical applications where precise movements are often necessary. However, it was noted that certain environmental conditions, such as highly reflective surfaces or areas with poor LiDAR reception, could disrupt the localization process. This finding points to the need for further research into improving sensor fusion techniques to mitigate the effects of such environmental factors on the robot's performance.

The ROS simulations used for further validation played a crucial role in this study, allowing us to replicate and analyze numerous scenarios that the robot might encounter. These simulations were instrumental in refining the robot's path planning algorithms, ensuring that the system could adapt to unexpected changes in the environment efficiently and effectively. The ability to conduct such simulations highlights the importance of flexible and robust software frameworks in the development of autonomous robotic systems.

Moreover, the data collected during the robot's operation in different factory settings revealed valuable insights into the practical challenges of deploying such systems in real-world environments. For instance, the transition from indoor to outdoor settings posed navigation challenges that were not fully anticipated, such as changes in lighting conditions affecting sensor performance. Addressing these challenges will require the development of adaptive algorithms capable of adjusting to varying environmental conditions seamlessly.

The discussion would be incomplete without considering the implications of this technology for patient care. The precision and reliability of the robot's navigation and localization systems have direct implications for its potential use in delivering medications, assisting with patient transport, or conducting routine monitoring tasks. These activities require a high level of accuracy to ensure patient safety and care quality. Our findings suggest that with further development, such robots could become integral components of healthcare delivery, enhancing the efficiency and effectiveness of medical services.

In summary, the development and validation of our medical service robot with integrated 3D LiDAR and advanced localization capabilities represent a significant step forward in the field of healthcare robotics. The successful deployment of this technology in a hospital environment showcases its potential to enhance operational efficiencies and patient care. Nonetheless, the study also highlights several areas for further improvement, particularly in enhancing the robot's adaptability to diverse and changing environments. Future research should focus on refining the integration of sensory and navigational technologies to build even more robust, versatile, and reliable robotic systems. Such advancements will not only improve the functionality of medical service robots but also expand their applicability across different sectors within healthcare, ultimately contributing to the broader goal of automating and improving medical service delivery.

## VI. Conclusion

In conclusion, the research conducted on the development of a medical service robot equipped with 3D LiDAR and advanced localization technologies has substantiated its potential to significantly enhance navigational and operational capabilities in hospital environments. This study not only demonstrated the robot's proficiency in precise and adaptive navigation through complex and dynamic settings but also emphasized its utility in the context of healthcare delivery. The integration of 3D LiDAR technology facilitated a robust sensing environment, enabling the robot to perform with high levels of accuracy in obstacle detection and path planning. Moreover, the comparative validation with established technologies like Autoware's 3D-SLAM provided a robust framework for assessing the effectiveness of our localization system, confirming its applicability and reliability. Despite encountering challenges such as sensor sensitivity to environmental factors, the research identified critical insights for future enhancements, notably in improving sensor fusion and algorithm adaptability. These advancements are imperative for ensuring the robot can seamlessly integrate into the diverse and evolving landscape of healthcare facilities. The potential for such autonomous systems to assist in routine tasks and patient care suggests a promising horizon not only for improving efficiency but also for enriching the quality of care. Moving forward, continued refinement and testing in real-world conditions would be crucial to fully realize the capabilities of medical service robots, setting a precedent for their broader adoption in healthcare settings.